Antonio San Martín & Catherine Trekker

# INTRODUCING HUMAN-CENTEREDNESS IN AI-ASSISTED LEXICOGRAPHY

**Abstract** This paper proposes a human-centered artificial intelligence (HCAI) framework for AI-assisted lexicography. While generative AI offers significant opportunities to enhance lexicographic work, it also raises concerns regarding the future role of lexicographers and the preservation of linguistic and cultural diversity. Drawing on HCAI principles and previous applications in other language professions, the paper identifies four interrelated dimensions through which AI integration in lexicography can be understood and critically examined: the augmented lexicographer, the sociotechnical context of AI integration, bias, and the design of AI-powered lexicographic tools. The framework argues that AI should augment rather than replace lexicographers, combining automation with meaningful human control. It further emphasizes the importance of preserving professional agency, mitigating AI-generated biases, and designing tools around the needs of lexicographers. By doing so, the paper provides a foundation for future research and the beneficial integration of AI into lexicographic workflows.

## 1 Introduction

The widespread availability of generative artificial intelligence (GenAI), particularly large language models (LLMs) such as those powering ChatGPT and Claude, has the potential to transform lexicographic practice (Fuertes-Olivera & Tarp, 2025; Lew, 2024). On the one hand, the growing ability of LLMs to provide lexical information has raised concerns about the future role of lexicographers and the continued relevance of dictionaries and other lexicographic products (de Schryver, 2023). On the other hand, artificial intelligence (AI) offers powerful new opportunities to support lexicographic work. These developments place lexicography at a critical juncture and raise a fundamental question: how can AI be integrated so that it enhances rather than undermines the work of lexicographers?

In this context, the paradigm of human-centered AI (HCAI) (Capel & Brereton, 2023; Schmager et al., 2023; Shneiderman, 2020) provides a useful framework for guiding the integration of AI into lexicography. HCAI places human needs and well-being at the center of AI development and challenges approaches that prioritize machine performance and cost reduction at the expense of human values. Applied to lexicography, this perspective recognizes that lexicographic resources provide important social benefits by supporting reliable access to lexical knowledge and preserving linguistic and cultural diversity. To protect these benefits, lexicographers must retain meaningful control over AI-assisted workflows, so that automation serves to augment professional expertise rather than displace or marginalize it.

Against this background, this paper proposes a conceptual framework for AI-assisted lexicography grounded in HCAI.[1] It identifies a set of interrelated dimensions through which the relationship between AI and lexicographic work can be understood and critically examined.

[1] This paper focuses on the application of HCAI to lexicographic work from the perspective of the lexicographer. Although the same approach can also be applied to the impact of AI on end users of lexicographic products, the present discussion concentrates on the former while acknowledging its direct implications for the latter.

By doing so, the paper seeks to provide a conceptual starting point for future research and support the beneficial integration of AI into lexicographic practice.

The remainder of the paper is organized as follows. Section 2 introduces the principles of human-centered AI and their application to language professions. Section 3 develops a human-centered framework for AI-assisted lexicography organized around four interrelated dimensions: the augmented lexicographer, the sociotechnical context of AI integration, bias, and the design of AI-powered lexicographic tools. Section 4 concludes by reflecting on the implications of the framework and identifying priorities for future research.

## 2 Human-Centered Artificial Intelligence

Much contemporary AI development prioritizes speed and efficiency over human values and societal impacts (Schmager et al., 2023, p. 1). In response, HCAI defines human flourishing as the central aim of AI (Vallor, 2024a, p. 13) and calls for equitable outcomes and broadly shared benefits for humanity (Stanford Institute for Human-Centered AI, 2025). It reconceptualizes AI as a tool designed to serve human needs and challenges the assumption that progress necessarily demands minimizing human involvement through automation. Instead, HCAI frames AI as amplifying human capabilities while preserving human control.

This reorientation also requires acknowledging that AI functions within sociotechnical ecosystems of people, institutions, and norms (Dignum & Dignum, 2020, p. 2). In profit-driven environments, organizations may impose AI to cut costs rather than empower users. HCAI also addresses ethical challenges in AI deployment (including bias, environmental impact, or intellectual property concerns) as well as the design of AI tools that prioritizes users' needs and goals.

Translation was the first language profession to which HCAI was applied (Jiménez-Crespo, 2025a; O'Brien, 2024). Although lexicography faces distinct challenges, translation offers a useful cautionary example. Neural machine translation has disrupted the field for over a decade, reshaping workflows and labor conditions. Organizations have often imposed AI tools to maximize speed and reduce costs at the expense of translators' needs. Moorkens (2020) describes this trend as "*digital Taylorism*," in which automation reduces translators to "*cogs in a large machine*," diminishing their agency. An HCAI approach to translation seeks to retain the benefits of automation while reducing its harms.

Among language professions, terminology work arguably faces AI-related challenges that most closely resemble those of lexicography, as both involve the systematic extraction and representation of lexical knowledge. This paper adapts the HCAI framework proposed for terminology (San Martín, 2025) to the specific context of lexicographic work.

## 3 Human-Centered AI-Assisted Lexicography

GenAI is arguably the most significant development in lexicography since the introduction of corpus analysis tools. For users, it enables new ways of accessing lexicographic information; for lexicographers, it offers the potential for greater efficiency. At the same time, however, it raises concerns about the future of lexicography, lexicographers, and lexicographic products (de Schryver, 2023).

GenAI chatbots are increasingly capable of fulfilling users' lexicographic needs and, in some respects, offer advantages over traditional dictionaries. They provide access to lexical information in a flexible and user-friendly manner, allowing users to formulate queries in context, receive customized answers, and ask follow-up questions.

However, GenAI-based chatbots are also prone to errors and hallucinations, may reproduce biases, and, due to their non-deterministic nature, generate inconsistent answers. These limitations raise important concerns about their reliability. In contrast, dictionaries produced through established lexicographic processes provide information that has undergone expert validation. Moreover, dictionaries are not merely tools for accessing lexical data. As Arias-Arias et al. (2025, p. 252) argue, they also hold cultural, social, and emotional value, functioning as "*repositories of cultural heritage and linguistic diversity, reflecting the evolution of language and society. They serve as educational artifacts, historical records, and symbols of linguistic identity*." In the case of prescriptive dictionaries, they additionally provide guidance on recommended usage.

From a human-centered perspective, the wholesale replacement of lexicography with GenAI would conflict with broader social values and benefits. Yet cost pressures and growing overtrust in AI risk marginalizing lexicographic expertise. AI will likely transform lexicography, which had already extended beyond dictionary compilation before GenAI emerged (Tarp, 2019, p. 231). Although some traditional dictionaries, particularly online ones, will probably survive, lexicographic authority may increasingly become embedded within broader AI-mediated language infrastructures. In response, Sennrich & Ahmadi (2025) propose the term "*conversational lexicography*" to describe an approach in which AI chatbots serve as interfaces for lexicographic content curated by lexicographers, combining the reliability of expert-curated content with conversational flexibility and user-friendliness.

HCAI envisions AI as augmenting rather than replacing lexicographers. In this view, GenAI serves as a powerful tool whose assistance can be broadly grouped into two overlapping modes: post-editing lexicography and intelligent support. Although post-editing lexicography (a term coined by Jakubíček et al., 2018) predates GenAI, recent advances enable unprecedented levels of automation, such as the automatic generation of definitions or lexicographic illustrations. Intelligent support, by contrast, includes tasks such as analyzing lexical data, assisting corpus analysis, and conducting quality assurance. Emerging agentic systems, including OpenAI's Codex and Anthropic's Claude Code and Cowork, further enable lexicographers to analyze and manipulate lexical data without requiring programming expertise.

Despite these powerful capabilities, GenAI has clear limitations as a lexicographic tool. Existing evaluations of LLMs in diverse lexicographic tasks[2] suggest that, while they can achieve strong results in certain contexts, particularly for high-resource languages, their performance remains imperfect and inconsistent. Furthermore, persistent biases continue to pose significant challenges.

As AI both threatens and transforms lexicography, a human-centered approach is essential to ensure that AI supports rather than marginalizes lexicographers. The following sections examine four interrelated dimensions of human-centered AI-assisted lexicography: the augmented lexicographer, the sociotechnical context of AI integration, bias, and the design of AI-powered tools.

[2] Due to space limitations, it is not possible to cite or review all relevant studies. Representative evaluations of LLMs include those by Chen et al. (2024), Kosem et al. (2024), and Rees & Lew (2024).

## 3.1 The Augmented Lexicographer

Technological tools have enhanced lexicographers' capabilities for decades. This computational augmentation enables lexicographers to work more efficiently and potentially produce higher-quality output than would be possible without technological support. Since lexicography is now inseparable from technology, the contemporary lexicographer can be understood as an augmented lexicographer[3]. This transformation was already accelerating before the emergence of GenAI. Krek (2019), for instance, argued that lexicography had undergone a computational turning point driven by rapid advances in natural language processing and the growing availability of language data.

From a HCAI perspective, the augmentation of lexicographers through AI should be guided by the two principles formulated by Shneiderman (2020), which are discussed in the following two subsections. The final subsection examines how GenAI is transforming the role of lexicographers and the skills required to effectively leverage AI-based augmentation.

### 3.1.1 High automation and strong human control

The first HCAI principle holds that high levels of automation and human control are not mutually exclusive but can coexist productively. Traditional AI approaches often frame automation and human oversight as a zero-sum relationship, a view HCAI explicitly rejects (Shneiderman, 2020, p. 115). Accordingly, under the right circumstances, AI-assisted lexicographers are expected to outperform both fully automated systems and entirely manual approaches, since human expertise and computational efficiency operate synergistically (Capel & Brereton, 2023, p. 6). Such augmentation, however, requires a careful balance between machine and human intervention. Excessive automation may compromise the quality of lexicographic output, whereas excessive human involvement can unnecessarily limit the efficiency gains that AI systems may provide (Shneiderman, 2020, p. 116).

This balance between automation and human control must be determined at the level of individual subtasks rather than for an entire task. In definition generation, for example, the lexicographer may retain control by designing the prompt that guides the process, specifying elements such as the target audience, intended perspective, and stylistic constraints, and providing a relevant corpus. The LLM may then automatically generate a draft definition. Human control may resume during the review and validation stage, where the lexicographer evaluates and revises the output as needed. At this point, automation may once again support the process by assisting with corpus analysis, while the lexicographer retains final authority over the definition ultimately adopted.

The balance cannot be defined uniformly across all lexicographic contexts. It depends on project goals and LLM performance in different tasks, domains, and languages. Consequently, the extent and form of automation may differ according to factors such as the work setting (commercial, institutional, academic, etc.), the type of lexicography involved (general, bilingual, learner, etc.), the languages concerned, and even the individual preferences of the lexicographer.

Achieving this balance requires systematic research on the performance of AI systems for lexicography. This includes benchmarking LLMs across lexicographic tasks and subtasks to identify strengths and limitations. It also requires examining AI use in situated practice: how lexicographers interact with tools throughout their workflow and whether these interactions actually improve their capabilities, in a manner comparable to translation post-editing studies (Vieira, 2019). Bedard et al. (2026) show that AI use can generate cognitive fatigue in some work contexts, underscoring the need to identify forms of automation that provide clear benefits. Following O'Brien's (2024, p. 400) proposal for translation, research should also elicit lexicographers' views on the types of AI augmentation they consider most useful and on how they would like these systems to support their work. For example, since corpus evidence constitutes a central component of lexicographic practice (Hanks, 2012), lexicographers will likely value its integration into AI-assisted workflows.

[3] Adaptation of the notion of "augmented translator" (DePalma & Lommel, 2017).

### 3.1.2 Amplification rather than emulation

Shneiderman's (2020) second HCAI principle holds that AI should not aim to replicate human intelligence. Artificial and human intelligence are fundamentally different in nature, and treating AI as if it should mimic human cognition misconstrues both forms of intelligence. Accordingly, HCAI conceives AI not as a substitute for human intelligence, but as a tool that extends it.

This distinction is crucial in lexicography. AI systems excel at processing large datasets and detecting patterns at great speed (Ozmen Garibay et al., 2023, p. 422). LLMs, however, manipulate language without direct access to the realities to which it refers (Poibeau, 2025, p. 48). As Vallor (2024b, p. 24) argues, no dataset can capture the open-ended complexity of lived human experience. Lexicographic work therefore cannot be reduced to statistical pattern recognition alone. Beyond corpus evidence, lexicographers can draw on world knowledge, embodied experience, introspective linguistic competence, and direct engagement with language communities.

Within this framework, AI contributes to lexicography because it offers strengths humans lack. For instance, it can quickly identify candidate senses, synonyms, antonyms, or interlingual equivalents that are time-consuming to extract from corpora. However, producing and validating reliable lexical data ultimately requires expert lexicographic judgement and human understanding of language and the world.

### 3.1.3 The new role and skills of lexicographers

As Tiberius et al. (2024) note, lexicographers do more than edit dictionary entries. Their roles increasingly encompass project management, data management, communication with computational specialists and end users, fundraising, and public relations. The integration of AI is expected to further transform the profession by reshaping both the tasks that lexicographers perform and the competencies that define their professional value.

Drawing on AI literacy frameworks proposed by Annapureddy et al. (2025) and Zhang & Magerko (2025), lexicographers will need to acquire skills such as understanding how different AI models can support lexicographic tasks, selecting and using appropriate AI tools, crafting effective prompts, and recognizing the ethical and legal implications of AI use. They must also critically evaluate AI-generated outputs, identifying and correcting errors and potential biases.

Lexicographers must also develop sufficient technical expertise to retain control over the integration of AI. Tarp (2015, p. 219) already argued that lexicographers require adequate computational knowledge to guide programmers, rather than allowing them to shape lexicographic decisions. Similarly, lexicographers today need a solid understanding of AI to supervise its integration and ensure that implementation decisions align with lexicographic principles and objectives.

Finally, developing AI-related competencies must not undermine the fundamental expertise that defines the lexicographic profession, a concern similarly raised by Jiménez-Crespo (2025b, p. 279) in relation to translators. If the role of lexicographers is reduced to merely validating AI-generated output, the profession risks progressive de-skilling.

## 3.2 The Sociotechnical Context of AI Integration in Lexicography

AI integration in lexicography is more than a technical matter, since it always takes place within a broader social context including a variety of *"stakeholders, institutions, cultures, norms and spaces"* (Dignum & Dignum, 2020, p. 2). This sociotechnical environment involves multiple stakeholders whose roles vary across commercial, institutional, and academic settings, including companies, universities, publishers, public institutions, lexicographers, programmers, end users, and the wider language community. These actors collectively shape both the ways AI systems are integrated into lexicographic workflows and the consequences of that integration.

While the traditional AI goals of machine autonomy and performance are not inherently problematic, augmenting lexicographers through AI cannot be considered human-centered if it dehumanizes them by prioritizing profit extraction or cost reduction. As Kalluri (2020, p. 169) argues, human-centeredness ultimately depends on where power resides. In lexicography, AI integration should primarily support lexicographers and the broader social benefits of their work, rather than allowing institutional and economic interests to undermine them.

AI-related tensions intensify pre-existing structural problems in lexicography, since many dictionary projects were already operating under conditions of financial fragility (Tiberius et al., 2024, p. 19). In this context, from the perspective of sustainable lexicography (Colman, 2016; Fuertes-Olivera, 2024), AI augmentation may help preserve some lexicographic projects by reducing costs. However, the same economic logic may also encourage funders to maximize automation in ways that marginalize lexicographers. The difference lies in whether cost savings are achieved by supporting expert work or by removing expert authority from decisions that determine the reliability and societal contribution of the lexicographic resource.

The sociotechnical conditions under which AI is implemented also directly affect lexicographers' well-being and their ability to carry out their professional responsibilities. For AI integration to remain human-centered, lexicographers must retain meaningful agency over how AI is incorporated into their workflows. When AI systems are imposed without such control, lexicographers risk being reduced to mere validators of machine-generated output, with negative consequences for professional autonomy, motivation, job satisfaction, and stress levels (Oviatt, 2021, p. 278). More importantly, such conditions may limit lexicographers' capacity to produce reliable and socially valuable resources that reflect linguistic and cultural diversity.

In this context, the benchmarking and evaluation of LLMs acquire an additional human-centered function. Beyond measuring technical performance, such research can help identify which aspects of lexicographic expertise current AI systems cannot reliably reproduce. It can therefore provide further evidence that expert human judgment remains necessary for ensuring the quality and social value of lexicographic resources. In doing so, it also provides lexicographers with empirical grounds to defend their professional role and to claim meaningful authority over AI-assisted workflows.

## 3.3 Bias

Major sources of LLM bias relevant to lexicography include training data composition and algorithmic design (Ferrara, 2023). While the exact composition of the training data used by major proprietary LLMs remains undisclosed (Ravichander et al., 2025, p. 1962), the Common Crawl corpus is widely considered their largest component (Baack, 2024, p. 2199). This corpus consists largely of uncurated web data, over 40% of which is in English. Biases present in the training data are subsequently amplified by algorithmic processes. Because LLMs reproduce statistical patterns in their training data and are optimized to generate high-probability outputs, they reinforce and overrepresent dominant patterns (Sourati et al., 2026, p. 5). These factors produce distinct but often overlapping forms of bias, discussed separately below.

*Linguistic bias*. Linguistic bias denotes the tendency of LLMs to favor certain languages or linguistic features in their outputs. Since English is overrepresented in the training data of many leading LLMs, AI-generated lexicographic content may be more accurate in this language. Cross-linguistic interference may also occur, whereby one language, typically English, influences outputs in other languages, as shown by Rigouts Terryn and de Lhoneux (2024). Linguistic bias may also operate within a single language, when dominant regional varieties are projected onto less-resourced ones.

*Temporal bias*. Temporal bias refers to the risk that LLMs generate lexicographic information that is misaligned with the relevant time frame. Models may present diachronic data as synchronic, or vice versa, and may hallucinate when evidence is sparse. Although temporal bias in AI-assisted lexicography remains understudied, studies report LLM weaknesses in date-sensitive factual accuracy and chronology (Wallat et al., 2024). A key cause is that training corpora often mix historical and contemporary texts without clear temporal boundaries (B. Zhao et al., 2024, p. 15015). In the case of neology, temporal bias may further stem from the training data cutoff.

*Cultural bias*. Cultural bias denotes the tendency of LLMs to reproduce the cultural assumptions, values, norms, and perspectives that dominate their training data while underrepresenting or distorting others. Consequently, LLM outputs may reflect Anglophone and Western cultural interpretations as universal, a tendency reported by Tao et al. (2024).

*Epistemic bias*. Epistemic bias refers to the tendency of LLMs to overrepresent dominant forms of knowledge by amplifying what is most prevalent in their training data while marginalizing less represented perspectives. In practice, this often entails privileging mainstream conceptualizations prevalent in the Global North. This is particularly relevant for specialized language, since terms may be conceptualized differently across domains, theoretical traditions, and ideological currents (San Martín, 2022). Consequently, AI-generated lexicographic output may obscure conceptual variation, disciplinary disagreement, or emerging forms of knowledge.

*Social bias*. Social bias refers to the reproduction of prejudicial views, stereotypical representations, and discriminatory attitudes directed toward particular social groups on the basis of factors such as gender, race, ethnicity, age, sexual orientation, and disability (Gallegos et al., 2024, p. 1098). Although post-training techniques are intended to mitigate such biases, social bias remains an unresolved problem in LLMs (McIntosh et al., 2024).

The forms of AI bias discussed above have different consequences. Linguistic and temporal bias primarily produce descriptive distortions, since they compromise the faithful representation of language use. The role of the lexicographer is therefore to ensure descriptive adequacy by critically evaluating AI-generated output against corpus evidence and other reliable sources.

In prescriptive contexts, lexicographers must also prevent such distortions from being transformed into normative recommendations. This task specifically requires lexicographic expertise, since identifying subtle linguistic interferences and distortions cannot be fully automated. If left uncorrected, these forms of bias may contribute to the erosion of linguistic diversity.

Cultural, social, and epistemic bias raise a different type of problem because they concern visibility and legitimacy rather than descriptive accuracy alone. Although such biases in lexicography long predate AI (Kaplan, 2020), LLMs introduce new mechanisms for reproducing and amplifying them at scale. AI-generated lexicographic content may reproduce existing asymmetries while presenting them under the appearance of computational objectivity.

Critical lexicography has already emphasized that dictionaries are not neutral representations of language (W. Chen, 2019, p. 362), but rather products shaped by social, cultural, and ideological values. Given that, from the point of view of cognitive linguistics, meaning cannot be separated from encyclopedic knowledge (Evans, 2019, p. 386), representing language use necessarily involves selecting and organizing culturally and socially situated knowledge about the world. Consequently, even a strictly descriptivist approach cannot be fully neutral, as lexicographers must decide, in accordance with the aims and characteristics of a given project, which usages and perspectives to include and which to background or omit. Descriptivism, therefore, cannot be regarded as entirely neutral, because any description of language inevitably reflects particular social and epistemic standpoints.

From a human-centered approach, the role of lexicographers is therefore not to pursue an unattainable neutrality, but to critically manage AI-generated lexical representations through fairness and contextually adequate pluralism. This responsibility requires expert lexicographic judgment that cannot be delegated entirely to AI systems. Without the intervention of a lexicographer, AI outputs may privilege dominant cultural perspectives while distorting and marginalizing others, suppress epistemic diversity, and reinforce discriminatory representations affecting already marginalized social groups.

Managing AI-generated bias in lexicography requires intervention at multiple stages of the workflow. Upstream, lexicographers can reduce bias through careful prompt design and by constraining retrieval-augmented generation (RAG)[4] to appropriate sources. Downstream, bias mitigation requires the critical evaluation of AI-generated output through expert judgment, corpus analysis, and, in some cases, even the strategic use of AI systems themselves as bias-detection aids.

If lexicographers do not retain meaningful control over the integration of AI into lexicographic workflows, inaccurate and biased representations of the lexicon may circulate and acquire unwarranted authority. Such lexicographic information may subsequently influence real language use. The problem becomes even more significant when the resulting texts enter corpora used in lexicographic work or are incorporated into the training data of LLMs. This may contribute to self-reinforcing feedback loops that progressively amplify linguistic distortions as well as cultural, epistemic, and social inequities.

Research can contribute to bias mitigation by systematically evaluating LLMs across different types of bias, lexicographic tasks, and linguistic contexts. Such research would enable lexicographers to anticipate systematic distortions, make more informed methodological decisions, and maintain effective critical oversight over AI-assisted lexicographic workflows.

4 RAG is a mechanism that allows LLMs to base their responses on external information.

### 3.4 The Design of AI-Powered Lexicographic Tools

Some publicly available tools used by lexicographers, such as TLex and AntConc, have begun integrating GenAI. However, the extent to which AI has been incorporated into in-house solutions, which are widely used in many lexicographic projects (Tiberius et al., 2024), is understudied. It is nevertheless reasonable to assume that much of the interaction between lexicographers and GenAI still occurs through standalone chatbots. From a human-centered perspective, this mode of interaction is often suboptimal. The need to move continuously between tools and chatbots introduces unnecessary friction into the workflow and limits the extent to which AI can meaningfully augment professional practice. Consequently, AI functionality should be integrated directly into lexicographic environments and workflows, provided that this does not undermine the lexicographer's autonomy. Alternatively, lexicographic tools can be integrated into AI agentic systems (e.g., Codex and Claude Code)[5]. Such integrations are not merely a technical issue but a design challenge that should be guided by established principles of human-computer interaction, particularly the well-developed field of human-centered design (HCD)[6], which addresses the physical, cognitive, social, and emotional factors of the human-machine relationship (Boy, 2011, p. 1).

Applying HCD principles (Benyon, 2019, p. 13) to AI-assisted lexicographic tools involves two fundamental requirements. First, systems and interfaces must be designed around the needs of lexicographers. Second, lexicographers themselves must be actively involved in the conception and refinement of AI-powered tools. This participatory approach helps avoid reverse adaptation (Winner, 2020, p. 174), whereby lexicographers and lexicographic projects are forced to adapt to technological constraints instead of having technology designed around lexicographic requirements. By placing lexicographers at the center of the design process, HCD also supports the practical implementation of broader HCAI principles, particularly the preservation of human control over AI systems and the genuine augmentation of lexicographers' capabilities.

With regard to preserving human control over AI, useful guidance can be drawn from recommendations developed in the context of HCAI-oriented translation research (Jiménez-Crespo, 2025a; O'Brien, 2024). These recommendations emphasize that AI-powered tools should combine strong usability with features that allow users to selectively enable or disable particular system functions and calibrate automation levels depending on the task or individual preferences. Role-aware permissions, version control, and the capacity of lexicographers to override AI suggestions could also ensure that they retain ultimate authority over lexicographic outputs.

Beyond preserving control, the interface of AI-assisted lexicographic tools should augment the lexicographer's capabilities by supporting decision-making. One important avenue for such augmentation is explainability, that is, the use of techniques that help users understand and evaluate AI outputs (Capel & Brereton, 2023, p. 6). Even though the internal processes of LLMs remain opaque (H. Zhao et al., 2024, p. 2), approaches such as RAG, chain-of-thought prompting, and reasoning models can provide, to a certain extent, evaluative cues about outputs.

Among these approaches, RAG is arguably one of the most valuable for lexicographic work because it enables AI outputs to be linked to identifiable and verifiable sources. The most desirable implementation involves retrieval from corpora selected by lexicographers themselves, since unrestricted web-based RAG may rely on unreliable, contextually inadequate, or even

5 Agentic systems combine the knowledge and reasoning capabilities of LLMs with the ability to write and execute code to operate and orchestrate external tools, such as corpus analysis tools and databases.

6 HCD has already been applied to lexicography from the perspective of the end users (Tarp & Gouws, 2020).

AI-generated material. This approach also has the advantage of pointing to specific linguistic evidence, a key issue in lexicography. Nonetheless, despite the value of RAG, it should not be the sole basis for LLM-generated suggestions, as doing so would limit the model's ability to draw on the broader knowledge acquired during training. LLMs can provide useful contributions that are not directly traceable to specific sources, but in such cases the relevant linguistic evidence must subsequently be identified and validated.

The augmentation of lexicographers also requires careful attention to the cognitive effects of automation. Two particularly important risks are reduced critical engagement and anchoring. Reduced critical engagement refers to a decline in the rigor with which users evaluate and validate AI-generated outputs (Heer, 2019, p. 1844). This tendency may be driven by fatigue, time constraints, or an automation-induced overtrust in the reliability of AI outputs that discourages verification (Klingbeil et al., 2024, p. 7). Tool design may mitigate this risk through empirically calibrated intentional friction, such as confirmation dialogs that encourage lexicographers to reassess AI suggestions before validation.

The anchoring effect may also emerge in AI-assisted lexicographic work when initial AI suggestions constrain subsequent human judgment and limit consideration of alternatives (Furnham & Boo, 2011). Early exposure to AI-generated proposals may thus narrow expert reasoning. One potential mitigation strategy is therefore to delay the presentation of AI suggestions until after lexicographers have completed an initial independent analysis. Such an approach allows AI to function as a source of support and comparison rather than as a starting point that shapes the entire decision-making process.

Taken together, these design recommendations should be regarded as provisional hypotheses rather than established best practices. Their effectiveness will ultimately depend on systematic empirical validation across a range of lexicographic contexts and workflows. Advancing AI-assisted lexicography therefore requires sustained collaboration between researchers, developers, and lexicographers, as well as the rigorous application of established HCI methodologies, including interviews, focus groups, usability studies, participatory design approaches, and observational research (Capel & Brereton, 2023, p. 4).

## 4 Conclusions

AI is likely to transform lexicography profoundly. The key question is not whether AI will be integrated into lexicographic work, but how. This paper has argued that HCAI provides a principled framework for ensuring that AI supports and strengthens lexicographic practice rather than undermines it. To this end, the paper has proposed a human-centered framework for AI-assisted lexicography organized around four interrelated dimensions: the augmented lexicographer, the sociotechnical context of AI integration, bias, and the design of AI-powered lexicographic tools.

A central premise of this orientation is that lexicographers should remain at the core of AI-assisted lexicography. From a human-centered perspective, AI should augment rather than replace the lexicographer's capabilities. High levels of automation and meaningful human control are not mutually exclusive; instead, they can coexist synergistically when AI systems are designed around the needs of lexicographers, preserving their decision-making authority and professional agency.

The implications of this framework extend well beyond lexicography as a profession, since lexicographic resources shape and support human communication. Preserving lexicographic expertise therefore serves broader societal goals. Decisions made now about AI deployment will influence not only professional practice but also the long-term preservation of linguistic and cultural diversity.

Advancing a human-centered approach to AI-assisted lexicography requires a sustained empirical research agenda. Future studies should investigate the performance of LLMs across lexicographic tasks. They should also examine how lexicographers interact with AI in situated practice and determine which forms of automation genuinely augment their capabilities. Further research should also assess the impact of AI on lexicographers' working conditions and autonomy, evaluate the forms of bias that emerge in AI-assisted workflows in order to identify strategies to manage them, and compare alternative design approaches to integrating AI into the tools used by lexicographers. The framework proposed in this paper serves as a starting point for research aimed at ensuring that AI empowers lexicographers to continue contributing to human flourishing in an increasingly AI-mediated society.

## Contact information

Antonio San Martín
University of Quebec in Trois-Rivières
antonio.san.martin.pizarro@uqtr.ca

Catherine Trekker
University of Quebec in Trois-Rivières
catherine.trekker-seguin@uqtr.ca

## Acknowledgements

This research was funded by Canada's Social Sciences and Humanities Research Council (grant number 430-2023-0248), Quebec Research Fund (grant 366841), and Spain's Ministry of Science and Innovation (grant number PID2020-118369GB-I00).